\documentclass[a4paper,UKenglish,cleveref, autoref,pdfa]{lipics-v2021}

\nolinenumbers

\usepackage{svg}
\usepackage[utf8]{inputenc}
\usepackage[T1]{fontenc}
\usepackage{caption}
\usepackage{xcolor}
\usepackage{listings}
\usepackage{pmboxdraw}

\hideLIPIcs  

\title{Towards Exploratory Reformulation of Constraint Models} 

\author{Ian Miguel}{School of Computer Science, University of St Andrews, UK}{ijm@st-andrews.ac.uk}{https://orcid.org/0000-0002-6930-2686}{EPSRC grant EP/V027182/1}
\author{Andr\'as Z. Salamon}{School of Computer Science, University of St Andrews, UK}{Andras.Salamon@st-andrews.ac.uk}{https://orcid.org/0000-0002-1415-9712}{}
\author{Christopher Stone}{School of Computer Science, University of St Andrews, UK}{cls29@st-andrews.ac.uk}{https://orcid.org/0000-0002-9512-9987}{EPSRC grant EP/V027182/1}

\authorrunning{Miguel et al.} 

\Copyright{Ian Miguel and András Z. Salamon and Christopher Stone} 

\ccsdesc[500]{Theory of computation~Constraint and logic programming}

\keywords{exploratory reformulation, constraint programming, Essence, graph rewriting} 

\category{} 

\relatedversion{} 

\EventLongTitle{PTHG-23: The Sixth Workshop on Progress Towards the Holy Grail}
\EventShortTitle{PTHG 2023}
\EventAcronym{PTHG}
\EventYear{2023}

\usepackage{xspace}

\newcommand{\code}[1]{{\lstinline!#1!}}
\newcommand{\essence}[0]{\textsc{Essence}\xspace}
\newcommand{\conjure}[0]{\textsc{Conjure}\xspace}
\newcommand{\eprime}[0]{\textsc{Essence Prime}\xspace}
\newcommand{\savilerow}[0]{\textsc{Savile Row}\xspace}

\usepackage{listings}
\begin{document}

\maketitle

\begin{abstract}
It is well established that formulating an effective constraint model of a problem of interest
is crucial to the efficiency with which it can subsequently be solved. Following from the observation that it is difficult, if not impossible, to know {\em a priori} which of a set of candidate models will perform best in practice, we envisage a system that explores the space of models through a process of reformulation from an initial model, guided by performance on a set of training instances from the problem class under consideration. We plan to situate this system in a refinement-based approach, where a user writes a constraint specification describing a problem above the level of abstraction at which many modelling decisions are made. In this position paper we set out our plan for an exploratory reformulation system, and discuss progress made so far.
\end{abstract}

\section{Introduction}

It is well established that formulating an effective constraint model of a problem of interest is crucial to the efficiency with which it can subsequently be solved \cite{freuder2018:progress}. This has motivated a variety of approaches to automating the modelling process. Some
learn models from, variously, natural language~\cite{Kiziltan2016:constraint}, positive or negative examples~\cite{DeRaedt2018:learning,Bessiere2017:constraint,Arcangioli2016:multiple}, membership queries, equivalence queries, partial queries~\cite{Beldiceanu2012:model, Bessiere2013:constraint}, generalisation queries~\cite{Bessiere2014:boosting} or arguments~\cite{Shchekotykhin2009:argumentation}. 
Other approaches include: automated transformation of medium-level solver-independent constraint models \cite{Rendl2010:thesis,Nethercote2007:minizinc,OPLBook,Mills1999:eacl,Nightingale2014:automatically,savilerow,Nightingale2015:automatically}; deriving implied constraints from a constraint model \cite{frisch2003:cgrass,colton2001:constraint, charnley2006:automatic, Bessiere2007:learning,Leo2013:globalizing}; case-based reasoning \cite{Little2003:using}; and refinement of abstract constraint specifications \cite{Frisch2005:rules} in languages such as ESRA \cite{Flener2003:esra}, \essence \cite{frisch2008:essence}, ${\mathcal F}$ \cite{Hnich2003:function} or Zinc \cite{marriott2008:design,ZincModref10,Rafeh2016:linzinc}.

Following from the observation that it is difficult, if not impossible, to know {\em a priori} which of a set of candidate models will perform best in practice, we envisage a system that {\em explores} the space of models through a process of reformulation from an initial model, guided by performance on a set of training instances from the problem class under consideration.

We plan to situate this system in a refinement-based approach, where a user writes a constraint specification describing a problem above the level of abstraction at which many modelling decisions are made. The advantage of proceeding from a problem specification rather than a concrete constraint model are that the structure apparent in a concise abstract specification, which may be obscured in concrete model, can help to guide reformulation. Furthermore, a single reformulated specification can be refined into a variety of both models and solving paradigms, allowing us to gain a fuller picture of performance.

In the remainder of this position paper we set out our plan for an exploratory reformulation system, and discuss progress made so far.

\section{Background: Essence and the Constraint Modelling Pipeline}

The refinement-based approach in which we intend to implement exploratory reformulation is the constraint modelling pipeline that takes an abstract problem specification in \essence~\cite{frisch2008:essence} as its input.
\essence is a well-established declarative language for constraint programming, supported by the Athanor local search solver~\cite{Attieh2019:Athanor} and the Conjure~\cite{Akgun2022:Conjure} and Savile Row~\cite{savilerow} translators working in concert with many back-end solvers such as the SAT solvers Cadical and Kissat~\cite{cadical} or the constraint solver Minion~\cite{Gent2006:minion}.
In this section we give a brief overview of the process of producing constraint models from \essence input.

An illustrative \essence specification of the Progressive Party Problem (problem 13 at \href{https://www.csplib.org/}{CSPLib}) is presented in \cref{fig:PPP}. The natural language description of the problem, taken from CSPLib~\cite{csplib:url}, is:

\begin{quote}
\em
The problem is to timetable a party at a yacht club. Certain boats are to be designated hosts, and the crews of the remaining boats in turn visit the host boats for several successive half-hour periods. The crew of a host boat remains on board to act as hosts while the crew of a guest boat together visits several hosts. Every boat can only hold a limited number of people at a time (its capacity) and crew sizes are different. The total number of people aboard a boat, including the host crew and guest crews, must not exceed the capacity. A guest boat cannot not revisit a host and guest crews cannot meet more than once. The problem facing the rally organizer is that of minimizing the number of host boats.
\end{quote}

An \essence specification identifies: the input parameters of the problem class (\code{given}), whose values define an instance; optional further constraints on allowed parameter values (\code{where}); the combinatorial objects to be found (\code{find}); and the constraints the objects must satisfy (\code{such that}). An objective function may be specified (\code{minimising} in the example) and identifiers declared (\code{letting}). \essence supports a number of abstract type constructors, such as relations, functions, sequences, sets, and partitions. These may be arbitrarily nested, such as the set of functions that represents the schedule in the example.

\begin{figure}
    \centering
\begin{lstlisting}
given n_boats, n_periods : int(1..)
letting Boat be domain int(1..n_boats)
given capacity, crew : function (total) Boat --> int(1..)
 
find hosts : set of Boat,
     sched : set (size n_periods) of function (total) Boat --> Boat
minimising |hosts|
 
such that
   $ Hosts remain the same throughout the schedule
   forAll p in sched . range(p) subsetEq hosts,
   $ Hosts stay on their own boat
   forAll p in sched . forAll h in hosts . p(h) = h,
   $ Hosts have the capacity to support the visiting crews
   forAll p in sched . forAll h in hosts .
      (sum b in preImage(p,h) . crew(b)) <= capacity(h),
   $ No two crews are at the same party more than once
   forAll b1,b2 : Boat . b1 < b2 ->
      (sum p in sched . toInt(p(b1) = p(b2))) <= 1
\end{lstlisting}
\caption{\label{fig:PPP} \essence Specification of the Progressive Party Problem. }
\end{figure}

The abstract decision variables supported by \essence are not typically supported directly by solvers, and so an \essence specification must be {\em refined} via the automated modelling tool \conjure~\cite{Akgun2022:Conjure} into the generic constraint modelling language \eprime~\cite{sr-manual}.
There are generally many different refinement pathways, depending on decisions as to how to represent the decision variables and the constraints on them, whether and how to break symmetry, whether to channel between different representations, and so on, each leading to a different constraint model. \conjure features various heuristics so as to select a good model automatically. The \eprime
model is then prepared for input to a particular solver by \savilerow~\cite{savilerow}. Depending on the target, e.g.~SAT vs CP vs SMT, further modelling decisions are required at this stage.

\section{Exploratory Reformulation of Essence Specifications}

Even though both \conjure and \savilerow feature heuristics to refine a high quality model for a target solver, the refinement process is heavily influenced by the \essence specification from which refinement proceeds. By reformulating the specification we can open up new refinement possibilities and therefore new models. 

The reformulations we envisage include the transformation of the logical and arithmetic expressions in the specification, which will affect how it is refined to a constraint model and encoded for a solver. Furthermore, by choosing to reformulate at the \essence level rather than a constraint model, we can take advantage of the structural information present in the abstract types that \essence provides.

To illustrate, we present a simple example reformulation of the Progressive Party Problem specification. The decision variable \code{sched} is a fixed-cardinality (for the number of periods) set of total functions. For each such function we might consider if we can further constrain its domain and range. Since the functions are total their domain is fixed to the set of boats. The range of each function has size at least one, since all boats have an image, and at most \code{n_boats} if these images are distinct.

The constraint:
\begin{lstlisting}
$ Hosts remain the same throughout the schedule
forAll p in sched . range(p) subsetEq hosts,
\end{lstlisting}
connects the size of the range of each function to that of the hosts set. Since \code{range(p)} is a subset of \code{hosts} and we showed above that \code{range(p)} has size at least one, \code{hosts} cannot be empty, so we can strengthen the \code{find} statement:
\begin{lstlisting}
find hosts : set (minSize 1) of Boat
\end{lstlisting}

From the above and the constraint:
\begin{lstlisting}
forAll p in sched . forAll h in hosts . p(h) = h
\end{lstlisting}
we can prove that \code{range(p) = hosts}, strengthening the first of the constraints in the specification: since \code{hosts} is a set, the \code{h} in the image of the function are distinct. So, \code{forAll h in hosts . p(h) = h} tells us that the range is at least the size of \code{hosts}. We showed that the range is a subset of the hosts, and it is the same size, so they are equal.

We could go further still to realise that each total function is in fact partitioning the boats into \code{|hosts|} parts, and reformulate the decision variable as follows:
\begin{lstlisting}
find sched : set (size n_periods) of partition from Boat
\end{lstlisting}

In general, for any formulas $a(x)$ and $b(x)$ in which the variable $x$ appears free, from the constraint \code{forall h in hosts . (a(h) < b(h))} we can derive the implied constraint \code{(sum h in hosts . a(h)) < (sum h in hosts . b(h))}.
Hence, from:
\begin{lstlisting}
$ Hosts have the capacity to support the visiting crews
forAll p in sched . forAll h in hosts .
   (sum b in preImage(p,h) . crew(b)) <= capacity(h),
\end{lstlisting}
we can derive that the sum of crew sizes is less than or equal to sum of host capacities:
\begin{lstlisting}
(sum h in hosts . (sum b in preImage(p,h) . crew(b)))
   < (sum h in hosts . capacity(h))
\end{lstlisting}

This collection of relatively simple reformulations is indicative of those we intend to build into our system. The hope is that combinations of simple reformulations can, in aggregate, make a significant improvement to an input specification. Of course, it is difficult to know whether a particular reformulation, or sequence of reformulations, will have a positive effect. In the above example, would we expect the total function or partition specification to perform better? The answer is that it depends exactly how these abstract structures are refined and encoded.

This motivates the need for exploratory reformulation, in which we consider the performance of different sequences of reformulations on a set of training instances for the problem class studied. Our proposal can also be seen as an extension of \cite{Akgun2021:towards}, which described heuristics to guide the choice of rewrite rules to apply to an {\sc Essence} specification. The focus in that work was \emph{type strengthening}, whereby properties expressed by constraints can sometimes be expressed instead by additional type information, allowing more effective model refinement.

In a more exploratory setting, it is unlikely that an exhaustive search through all possible reformulations will be possible. In order to control the exploration of new reformulation sequences versus the exploitation of existing sequences by extending them, we plan to employ Monte Carlo Tree Search, as has recently proven successful in the generation of streamliner constraints \cite{SPRACKLEN2023103915}.
Given a resource budget, the system will then explore a number of promising reformulated specifications in an attempt to improve on the original. The cost of this process is then amortised over the remainder of the problem class.

\begin{figure}
  \centering
  \includegraphics[width=0.75\linewidth]{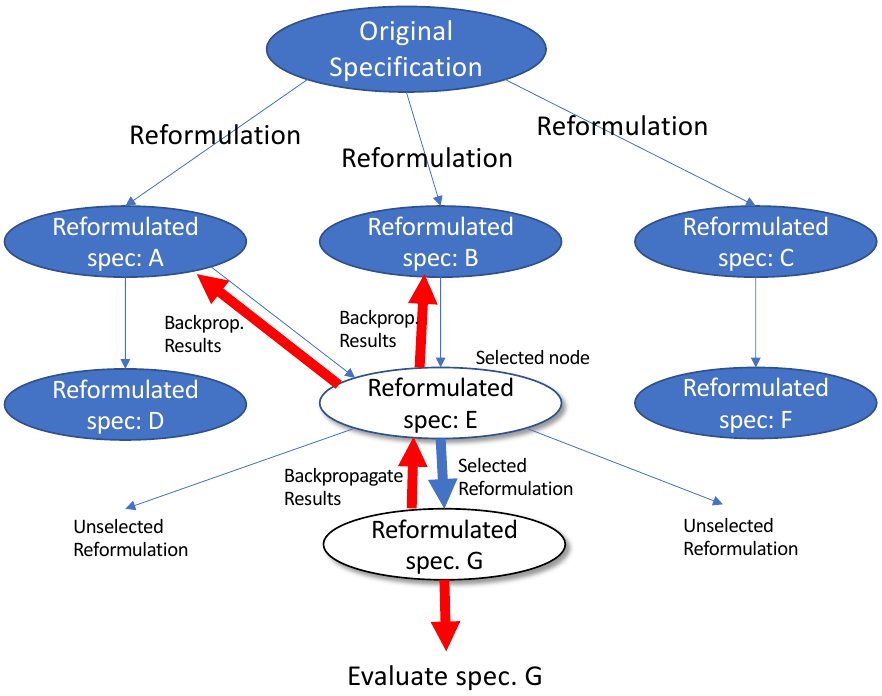}
  \caption{Exploratory Reformulation via Monte Carlo Tree Search.}
  \label{fig:MCTSReformulation}
\end{figure}

This process is illustrated in \cref{fig:MCTSReformulation}. The upper part of the tree of possible reformulations is maintained explicitly. Every iteration begins with a selection phase, which uses a policy such as Upper Confidence Bound applied to Trees (UCT) \cite{browne2012survey} to traverse the explored part of the tree until an unexpanded node is reached. The selected node is then expanded by randomly selecting a child, i.e.~a reformulation applicable to the specification represented by the selected node. The new reformulated specification is then evaluated against a set of training instances and the results back-propagated up the tree to influence the selection of the next node to expand.

In the remainder of this paper, we discuss our progress to date in implementing an exploratory reformulation framework for {\sc Essence} specifications.

\section{Current Progress}

\subsection{Reformulation, redux}

Reformulation can be conceptualised in many different ways.
Here we have chosen to pursue one particular framework for reformulation, based on rewriting.

The specification at each stage of reformulation can be represented as an abstract syntax tree (AST).
A specification can be recovered from the corresponding AST without loss of information.
ASTs are often modelled as trees.
It is also possible to replace common subtrees in the AST by pointers which turns the data structure into a form of directed acyclic graph.
Either way, ASTs can be represented as graphs consisting of a set of labelled vertices together with a set of directed and labelled arcs between vertices.
A \emph{node} in the AST is a labelled vertex.

A graph rewriting system nondeterministically matches a pattern graph to the target graph; if a match is found then the part of the target graph that is matched is replaced according to the rewrite rule by a different graph.
The details of matching and the kinds of rewriting can vary, but the graph rewriting paradigm is general enough to be Turing-complete~\cite{habel2001computational} and thus can capture the reformulation sequences that we want to study.

Each reformulation in a sequence of reformulations can be treated as a step taken by a graph rewriting system acting on the AST as the target graph being rewritten.
Each kind of reformulation is expressed by a graph rewrite rule.

We have used the graph rewriting language GP2~\cite{plump2017imperative} to perform rewriting on the abstract syntax tree representing a specification.
The GP2 system includes a flexible language in which to express graph rewriting rules, with an efficient implementation of the rewriting engine optimised for sparse large graphs~\cite{campbell2020improved}.

\subsection{System development}

Since \essence is a language with many features, we have initially defined a subset of the \essence language (Emini) that is sufficiently expressive to capture the full power of \essence itself, but with less syntax.
In particular, an Emini specification is also valid \essence.
Emini therefore inherits the decidability of satisfiability of \essence specifications.
The Emini language uses tuples and relations, integer and Boolean types, and allows most \essence expressions (including quantification, inequalities, Boolean logic, and arithmetic).
This choice of types was guided by previous work on the expressivity of \essence~\cite{Mitchell2008:expressive}, which showed that nested relation types are sufficient to express all problems in the polynomial hierarchy.
Over time, we intend to extend our system to the entire \essence language.

\begin{figure}
 %
\includegraphics[width=1\linewidth]{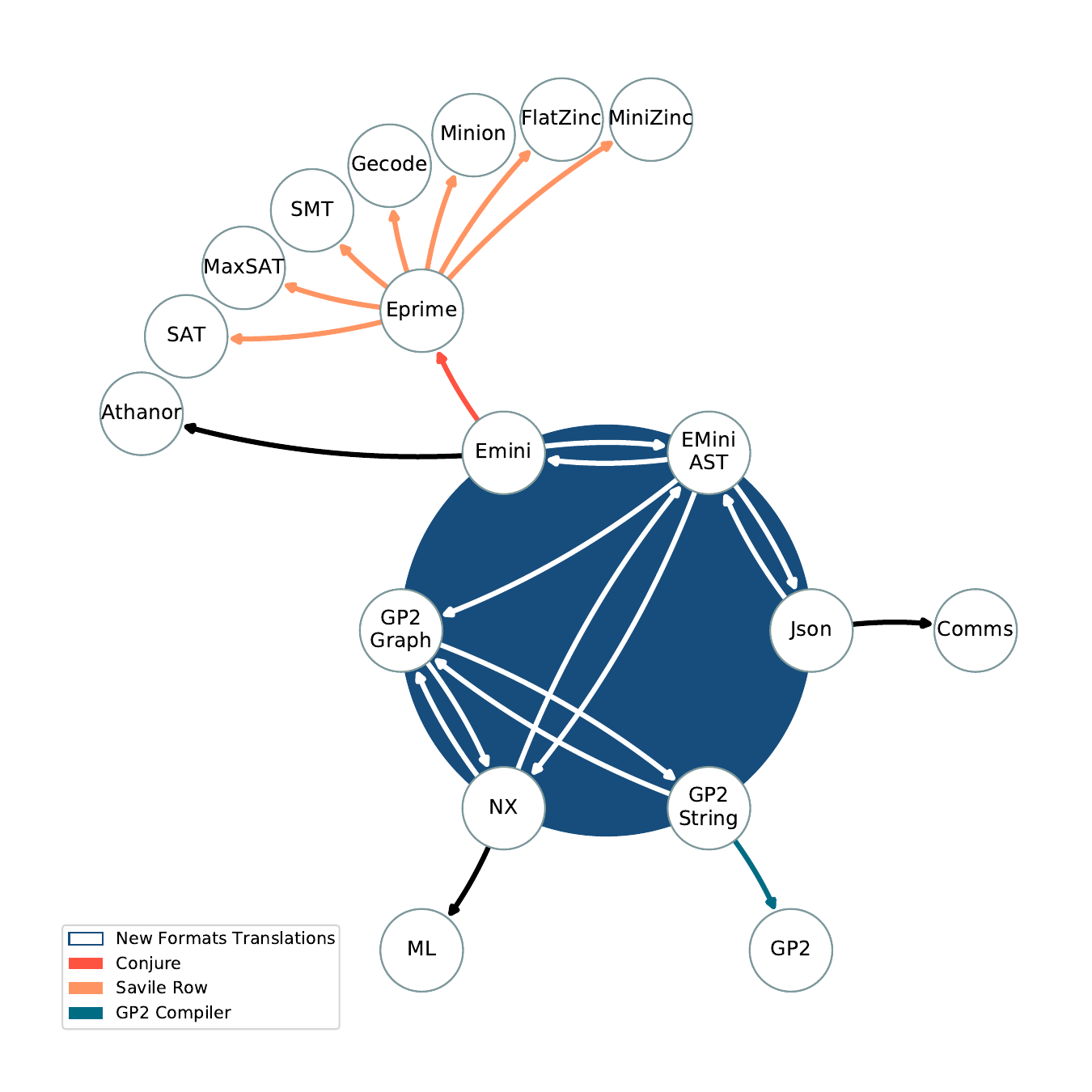}
\caption{Mapping all formats and transformations. The white arrows are all the novel translations that have been implemented. Abbreviations: ML=Machine learning, Comms=Communication, NX=NetworkX graph.}
\label{fig:TMAP}
\end{figure}

We have implemented our system in Python.
An AST representation of Emini is our core data structure. The AST is represented as nested Python objects and labels which are produced by a parser developed for Emini.
The AST can be translated to and from several other different formats.
Translations implemented to date are illustrated in \cref{fig:TMAP}.

Among these alternative formats, the NetworkX AST representation provides access to a variety of tools, such as plotting, graph algorithms, and machine learning libraries. The JSON format allows easy data interchange, and the GP2 format allows the application of graph rewriting rules to our specifications.

One application of our system is pretty printing.
Reading in an Emini specification, and then writing out an Emini representation of the AST, yields a specification in a normalised format with superfluous parentheses and syntactic sugar removed. Optionally, grammatical information about each node can be printed as in the following simple example specification:
\begin{lstlisting}[caption=Example specification, label=lst:simple]
find x : int(0..100)
such that
    1*(2+3)*4 = x
\end{lstlisting}
where we can visualise the AST of this specification as:
\begin{lstlisting}[ caption=AST, label=lst:AST, numbers=none, escapechar=!, basicstyle=\ttfamily, linewidth=0.82\linewidth,literate = {├}{{\textSFviii}}1 {─}{{\textSFx}}1 {└}{{\textSFii}}1 {│}{{\textSFxi}}1] 
└─ root  #Node
   ├─ find  #FindStatement
   │  └─ x  #DecisionVariable
   │     └─ int  #IntDomain
   │        ├─ 0  #Integer
   │        └─ 100  #Integer
   └─ such that  #SuchThatStatement
      └─ =  #BinaryExpression
         ├─ *  #BinaryExpression
         │  ├─ *  #BinaryExpression
         │  │  ├─ 1  #Integer
         │  │  └─ +  #BinaryExpression
         │  │     ├─ 2  #Integer
         │  │     └─ 3  #Integer
         │  └─ 4  #Integer
         └─ x  #ReferenceToDecisionVariable
\end{lstlisting}

\begin{figure}
    \includegraphics[width=\linewidth]{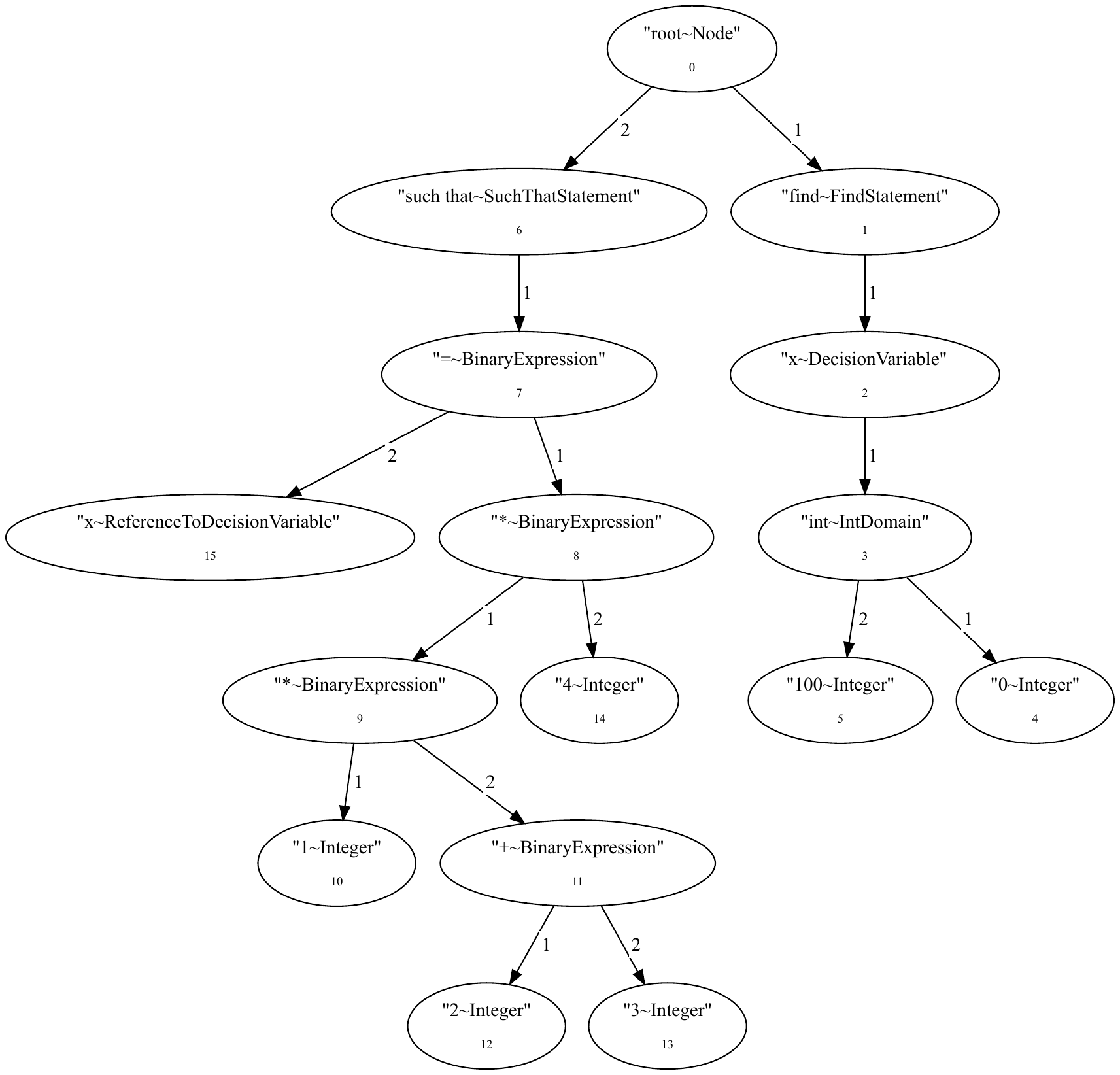}
  \caption{Visualisation of the GP2 graph representation.}
  \label{fig:gp2graph}
\end{figure}

We implement our rewrite rules as GP2 programs, and use the GP2 graph rewriting system to perform the rewriting nondeterministically.
The advantage of using GP2 is that this rewriting system performs well even on large graphs, and many different rewrite rules can be applied at once.
In the GP2 language graphs are specified by a list of vertices, represented by tuples (index, label) and a list of edges represented by tuples (index, source, target, label).
We store grammatical information in each node, and the parent-child relations become edges.
The ordering of a parent's children is represented by positive integers in the edge label, with 1 denoting the first child. The simple specification \cref{lst:simple}, with the AST in \cref{lst:AST}, is depicted in \cref{fig:gp2graph} as a GP2 graph.

\begin{figure}[ht]
\centering
\includegraphics[width=0.65\linewidth]{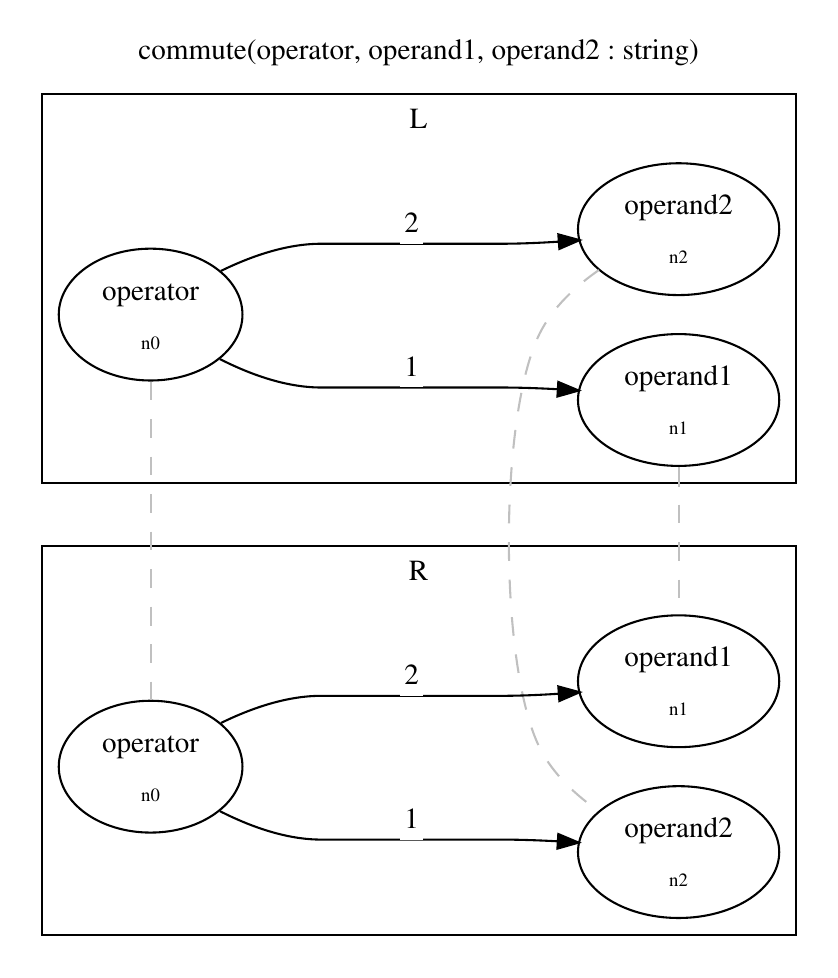}
\caption{Example GP2 rewrite rule that commutes the operands of a binary operator.}
\label{fig:gp2Program}
\end{figure}

The fundamental components of GP2 programs are rewrite rules.
Each rule is expressed as a pair of graphs that determine the precondition and postcondition of a matched subgraph.
\cref{fig:gp2Program} shows a simple example of a rule interchanging the operands of a commutative operation. Albeit trivial, this rewrite rule can already be used to test the behaviour of solvers over different variable orderings, and if equipped with an additional \textit{where} statement and a comparison, it could be used to normalise specifications.

We are currently investigating appropriate rewrite rules.

\section{Future Work}

The next steps in our road map are: commencing the production of instances for selected classes; automatically generate rewrite rules, starting with an initial set of hand-crafted ones; accumulating data on the effects and costs of reformulating the class specifications with a selected collection of rewrite rules; studying the ability of ML models to facilitate solving by selecting good rewrite rules.

Our use of the high-level \essence modelling paradigm enables automated generation of suitable benchmarks for each problem class, using our Generator Instance approach~\cite{Akgun2019:instance,Akgun2020:discriminating}.
This system automatically explores the space of valid instances based on the original problem specification expressed in \essence (or in our case, the Emini fragment).
The specification is transformed into a parameterised generator instance, and an automatic parameter tuning system is used to identify worthwhile regions of parameter space corresponding to instances of interest.

A key component of our system is the machine learning subsystem that selects rewrite rules. The aim is to select rules to apply to particular classes of problems, on a class level. The lattice of possible rewrite rule sequences is then explored using Monte Carlo tree search. We are currently implementing this aspect of the system.

\begin{figure}
\centering
\begin{subfigure}{.5\textwidth}
  \centering
  \captionsetup{justification=centering}
  \includegraphics[width=\linewidth]{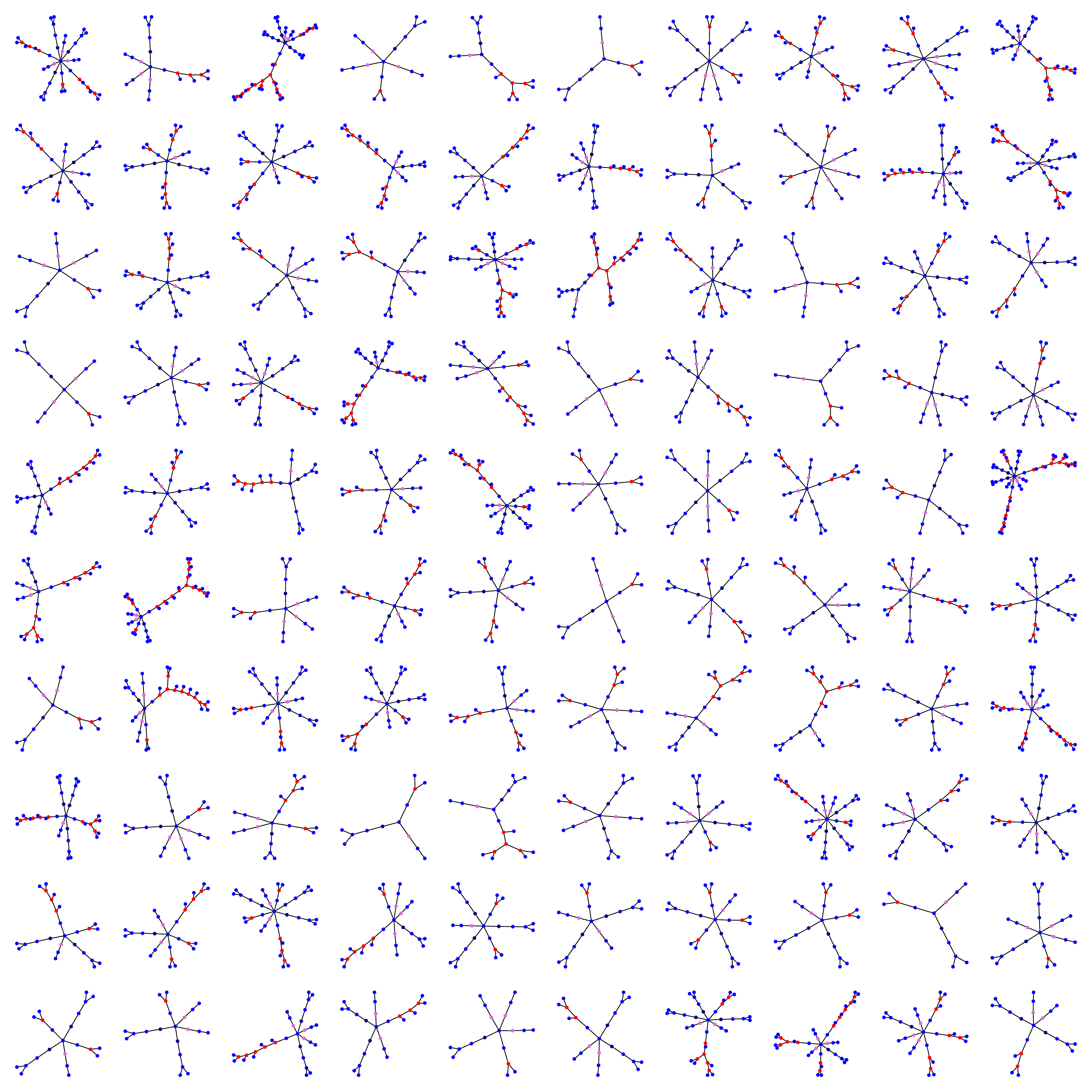}
    \caption{Abstract Syntax Trees}
    \label{fig:specsAST}
\end{subfigure}%
\begin{subfigure}{.5\textwidth}
  \centering
  \captionsetup{justification=centering}
  \includegraphics[width=\linewidth]{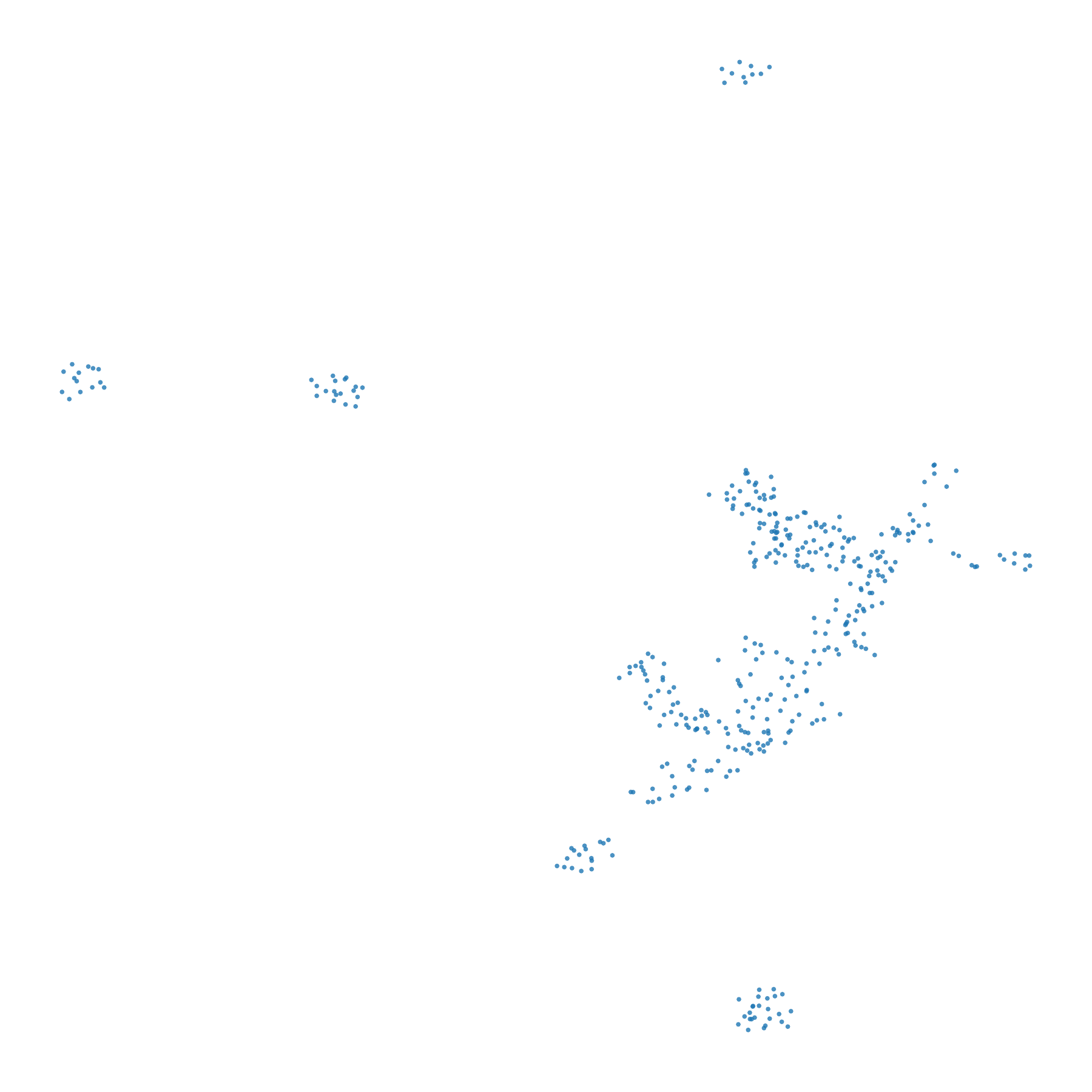}
    \caption{ASTs Embeddings}
    \label{fig:embedding}
\end{subfigure}
\caption{(a): A collection of different specifications in their abstract syntax trees form. Some key elements are highlighted. Red: binary expressions. Navy: decision variables. Pink: letting statements. (b): The ASTs of \cref{fig:specsAST} after embedding. Each dot is a specification, proximity of dots is due to structural similarities.}
\label{fig:test}
\end{figure}

One approach we are currently exploring is the use of graph embeddings which can isolate and identify particular structures in some vector space, making it amenable to further machine learning operations that work best, or exclusively, on tensor representations. In \cref{fig:specsAST} we show a collection of specifications, automatically produced with a hand-crafted generator for demonstrative purposes, displaying a variety of different structures. We turn the abstract syntax trees of those specifications into NetworkX graphs which are then embedded into a vector space using an unsupervised technique described in \cite{Rozemberczki2020:characteristic}. The results are shown in \cref{fig:embedding}. These types of embeddings, but even more so supervised ones that take into account the effect on performance of previously tested rewrite rules, can provide metrics that better inform which rewrite rules are likely to benefit a specific class, increasing the efficiency with which these are found.

One of the byproducts of these processes will be the creation of large amounts of semantically equivalent model variants.
This information will unlock an important component required to automatically learn and discover new metrics. Building on the idea that the distance between two models can capture their difference, and their proximity captures their sameness, we will provide the data and machinery capable of producing arbitrary amounts of distance zero examples. These components will also benefit those interested in studying the interactions between abstract specifications, reformulations, representations, and solvers.

We expect that improving our tools' ability to recognise that two models that appear different in structure and values are in fact the same or accurately estimating the magnitude of their difference, will enhance their ability to make better choices across the many stages of a problem solving pipeline.
These notions will be enriched by the data obtained from solving different reformulations that, even if semantically equivalent, can affect solvers in different ways.

A further virtue of our approach is that rewrite rules able to perform specification strengthening also enable the ability to synthesise altogether new specifications. This provides the ability to autonomously explore new problem classes.


\bibliography{main,general}

\end{document}